\documentclass[a4paper,fleqn]{cas-dc}

\usepackage[numbers]{natbib}
\usepackage{graphicx}
\usepackage{mathtools}
\usepackage{amsmath}
\usepackage{latexsym}
\usepackage{amssymb}
\usepackage{amsmath}
\usepackage{amsthm}
\usepackage{booktabs}
\usepackage{enumitem}
\usepackage{graphicx}
\usepackage{color}
\usepackage{grffile}
\usepackage{multirow}
\usepackage{amssymb}
\usepackage{pifont}
\usepackage{longtable}
\usepackage{xcolor}
\usepackage{tabularray}
\usepackage{tikz}
\usepackage{pifont}
\usepackage{setspace}
\definecolor{lightblue}{RGB}{27, 161, 226}
\definecolor{lightred}{RGB}{229, 20, 0}

\newcommand{\checkmarkicon}[2]{%
    \tikz[baseline=-0.5ex] \node[rectangle, fill=#1, inner sep=2pt, rounded corners=2pt] {\textcolor{#2}{\ding{51}}};%
}

\newcommand{\original}[1]{}

\newcommand{\xmark}{\ding{55}}%

\newcommand{\ourname}{\emph{InfoSel}}
\newcommand{\ournameplus}{\emph{InfoSel$^*$}}
\newcommand{\ournamelong}{\emph{Informed Selection}}
\newcommand{\printorcid}[1]{}

\def\tsc#1{\csdef{#1}{\textsc{\lowercase{#1}}\xspace}}
\tsc{WGM}
\tsc{QE}
\tsc{EP}
\tsc{PMS}
\tsc{BEC}
\tsc{DE}

\begin{document}
\let\WriteBookmarks\relax
\def\floatpagepagefraction{1}
\def\textpagefraction{.001}
\shorttitle{Black-box model ensembling}
\shortauthors{Y. Xia et~al.}

\title [mode = title]{Black-box model ensembling for textual and visual question answering via information fusion }                      



\author[1,2]{Yuxi Xia}\cormark[1]
\credit{Conceptualization, Data curation, Formal analysis, Investigation, Methodology, Software, Validation, Visualization, Writing– original draft, Writing– review \& editing}

\ead{yuxi.xia@univie.ac.at}


\affiliation[1]{organization={Faculty of Computer Science, University of Vienna},
                addressline={Kolingasse 14}, 
                city={Vienna},
                postcode={1090}, 
                country={Austria}}

\affiliation[2]{organization={UniVie Doctoral School Computer Science, University of Vienna},
                addressline={Kolingasse 14}, 
                city={Vienna},
                postcode={1090}, 
                country={Austria}}

\affiliation[3]{organization={Department of Computer Science, Aarhus University},
                addressline={Åbogade 34}, 
                city={Aarhus},
                postcode={8200}, 
                country={Denmark}}

\affiliation[4]{organization={Faculty of Philological and Cultural Studies, University of Vienna},
                addressline={Universitätsring 1}, 
                city={Vienna},
                postcode={1010}, 
                country={Austria}}
                
\author[3]{Klim Zaporojets}
\credit{Writing– review \& editing, Supervision}
\ead{klim@cs.au.dk}

\author[1,4]{Benjamin Roth}
\credit{Funding acquisition, Methodology, Writing– review \& editing, Supervision}
\ead{benjamin.roth@univie.ac.at}

\cortext[cor1]{Corresponding author}




\begin{abstract}
A diverse range of large language models (LLMs), e.g., ChatGPT, and visual question answering (VQA) models, e.g., BLIP, have been developed for solving textual and visual question answering tasks. However, fine-tuning these models is either difficult, as it requires access via APIs, rendering them as black-boxes, or costly due to the need of tuning a large number of parameters. To address this, we introduce \textbf{\ourname}, a data-efficient ensemble method that learns to dynamically pick the winner from existing black-box models for predictions on both textual and multimodal visual question answering tasks. Unlike traditional ensemble models, \textbf{\ourname} does not rely on prediction probabilities or confidences, which typically are not available in black-box models. 
Experimental results on four datasets demonstrate 
that our approach achieves an absolute increase of up to +5.19\% in the F1-score compared to standalone LLMs, up to +31.63\% in accuracy compared to VQA models, using only 1K training instances. Besides, \textbf{\ourname} surpasses baseline ensemble methods in data efficiency. We perform an in-depth analysis of the importance of fusing diverse multimodal information in ensembling.
\end{abstract}





\maketitle
\section{Introduction}

Large language models (LLMs) have demonstrated remarkable proficiency across a wide range of tasks, predominantly attributed to their ability to comprehend instructions and tap into vast repositories of high-quality data~\cite{bubeck2023sparks,laskar-etal-2023-systematic}. For example,  ChatGPT finds extensive utilization in daily textual question answering (TQA) tasks, rendering substantial convenience to a myriad of users~\cite{OpenAI2023GPT4TR}. Furthermore, for visual question answering (VQA) tasks, VQA models have exhibited exceptional versatility, primarily due to their capability to comprehend both visual and textual context~\cite{gong2023multimodal}. 

\original{However, recent work~\cite{Koco__2023,laskar-etal-2023-systematic} indicate that LLMs, such as ChatGPT, underperform on task-specific datasets.} 
However, recent work~\cite{Koco__2023, laskar-etal-2023-systematic} indicates that LLMs, such as ChatGPT, fall short of state-of-the-art performance on task-specific datasets such as question answering.
Similarly, VQA models ~\cite{li_blip_2022, li_unsupervised_2021,bao2022vlmo} face challenges when applied to specialized datasets due to the idiosyncrasies in the content, format or structure of these datasets~\cite{arora2018stronger}.
Unfortunately, fine-tuning
LLMs (e.g., LLaMA-2-70b-chat~\cite{touvron2023llama}) 
on task-specific data requires a large number of GPU hours. Alternatively, 
training smaller, task-specific models from scratch requires a large amount of labeled data to achieve comparable performance. 
Furthermore, fine-tuning LLMs through proprietary APIs with self-uploaded labeled training data not only requires LLM experts' knowledge but is also expensive.\footnote{\url{https://platform.openai.com/docs/guides/fine-tuning/}}
These fine-tuned models further remain black-box, with restricted access to details regarding architectural intricacies, model weights, training data, and even prediction confidences.

In order to address these computational and accessibility challenges associated with fine-tuning, we introduce a scalable ensemble method called \textbf{\ourname} (\ournamelong). \ourname~allows for training with just a few task-specific samples.
Unlike current LM-ensemble methods (e.g., MetaQA~\cite{puerto2021metaqa}) which depend on the confidence scores and thus can not be applied to black-box models like GPT3.5 text-davinci models, \ourname~does not rely on such information and offers black-box ensembling. 
Furthermore, our proposed ensemble method incorporates \textit{task-specific} optimization, allowing it to be easily adapted to different datasets, considering variations of both the inputs and predicted answers from the ensembled LLMs (\textit{base models}).
This contrasts with traditional ensemble methods such as OLA \cite{woods1997combination} and PageRank \cite{Brin2012ReprintOT}, 
which are not adapted to task-specific particularities (e.g., different features) of different datasets. 
Finally, our method efficiently deals with \textit{multimodal} inputs. 
Concretely, our results exhibit superior performance on multimodal VQA inputs compared to 
state-of-the-art PairRanker~\cite{jiang2023llmblender} ensemble method which is designed to work exclusively with text.
An in-depth exploration of the importance of utilizing diverse information as fused inputs to enhance the performance of ensemble training.



\begin{table*}[ht]
\centering
\renewcommand{\arraystretch}{1.25}
\caption{Our method (\ourname) aims to optimize 
\textbf{\textit{task-specific}}
ensembling of \textbf{\textit{black-box}} models, where confidences and parameters can not be accessed. We use only a small portion of training data (\textit{\textbf{data-efficient}}). 
We optimize the performance of the \textbf{\textit{ensemble}} instead of standalone fine-tuned (FT) models. 
Finally, our method is \textbf{\textit{multimodal}}, and applicable to VQA task. 
}
\begin{tabular}[h]{cc@{\hspace{0.2cm}}c@{\hspace{0.3cm}}c@{\hspace{.5cm}}c@{\hspace{.5cm}}c@{\hspace{.2cm}} }
\toprule

& Fine-tuning \cite{yosinski2014transferable} & Pair-Ranker \cite{jiang2023llmblender} & PageRank \cite{Brin2012ReprintOT} & OLA \cite{woods1997combination} & \textbf{\ourname~(ours)}  \\

\hline
{Task-specific} &   \checkmarkicon{lightblue}{white} & \checkmarkicon{lightblue}{white} &\xmark & \xmark & \checkmarkicon{lightblue}{white} \\
{Data-efficient} & \xmark & \xmark &\checkmarkicon{lightblue}{white}  &\checkmarkicon{lightblue}{white} &\checkmarkicon{lightblue}{white} \\

{Black-box} & \xmark  & \checkmarkicon{lightblue}{white} &\checkmarkicon{lightblue}{white} &\checkmarkicon{lightblue}{white} &\checkmarkicon{lightblue}{white}\\

Multimodal & \xmark &  \xmark & \checkmarkicon{lightblue}{white} &\checkmarkicon{lightblue}{white}  & \checkmarkicon{lightblue}{white}\\
Ensemble & \xmark  &  \checkmarkicon{lightblue}{white}& \checkmarkicon{lightblue}{white} &\checkmarkicon{lightblue}{white}  &\checkmarkicon{lightblue}{white}\\
\bottomrule

\end{tabular}

\label{tab:compare}
\end{table*}

\begin{figure*}[ht!]

     \centering
     \includegraphics[width=1.\textwidth]{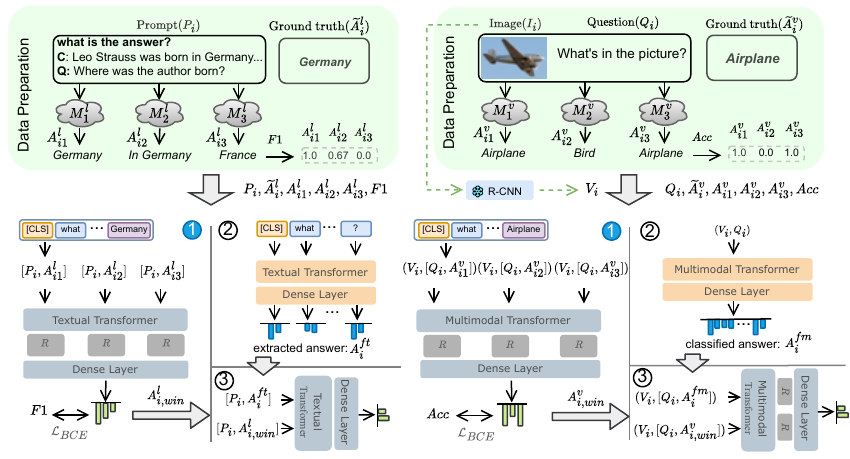}
   \caption{Architecture of our \textbf{\ourname} (step \raisebox{.5pt}{\textcircled{\raisebox{-.9pt} {1}}}) , FT (step \raisebox{.5pt}{\textcircled{\raisebox{-.9pt} {2}}})  and \textbf{\ournameplus} (step \raisebox{.5pt}{\textcircled{\raisebox{-.9pt} {3}}})  models.  $M^l_*$ and $M^v_*$ refer to black-box LLMs and VQA base models respectively, which are not trainable. The number of these base models is flexible, and is not restricted to 3 as in the figure. The models on the left (suffixed with -TT) are trained for the TQA tasks, while the models on the right (suffixed with -MT) are trained for the VQA tasks.
   All our models are trained independently. Note that FT and \ournameplus~ (step \raisebox{.5pt}{\textcircled{\raisebox{-.9pt} {2}}} and \raisebox{.5pt}{\textcircled{\raisebox{-.9pt} {3}}}) are optional 
   and are best suited for datasets that contain a high percentage of labels that the base models have not been exposed to.
   }
   \label{Fig:infosel}
\end{figure*}

At its core, \ourname~(see Figure \ref{Fig:infosel}) trains a small-size ensemble model to dynamically identify the most accurate base model (i.e., LLM or VQA model) for a given input, which we refer to as the \textit{winner}.
This is achieved by designing a meta-level classification task considering all the base models as labels for every input. We designed and implemented two ensemble architectures for textual and 
visual QA 
tasks. 
Our first proposed architecture, \ourname-TT, uses textual transformer (TT, 110M parameters)~\cite{devlin2019bert} as the backbone to generate a textual representation of the question with the predicted answers by base models.
Although \ourname-TT is straightforward and effective, it cannot handle multimodal data. To address this, we propose a second architecture named \ourname-MT, where we incorporate a multimodal transformer (MT, 115M parameters)~\cite{li2019visualbert} to generate fused contextual representations of a multimodal input (image, question, and the predicted answers).
These fused representations are used to train a dense layer to select the winner model. 
The challenge with this approach is the lack of exposure of the base models to new (unseen) labels appearing in the task-specific datasets. To address this, we fine-tune TT and MT models (FT-TT and FT-MT) separately to learn these new labels.
The predictions of these fine-tuned models are fused with the output from \ourname~using a second, separately trained \textbf{\ournameplus} ensemble model. We experiment with and without \ournameplus, as this component is considered optional when \ourname~is already performing well.


We select three LLMs (ChatGPT, LLaMA-2-70b-chat and GPT3.5 text-davinci-003) and three VQA models (ALBEF~\cite{li_align_2021}, BLIP~\cite{li_blip_2022} and VLMo~\cite{bao2022vlmo}). These models are used as ensemble base models to provide answers for textual and visual QA tasks respectively. 
 To demonstrate the \textit{data efficiency} 
 of the proposed architectures, we train them on a subsample of training data from public benchmark datasets and test on the corresponding full test data.
 %
 %
 %
Experimental results showcase improvements in the performance up to +5.19\% on 
textual QA
with \ourname \ and +31.63\% on 
VQA 
with \ournameplus \ when compared to the ensembled base models. 


In summary, our contributions are: 
\begin{itemize}
    \item \ourname, a novel data-efficient approach to ensemble black-box models without relying on access to model architecture, weights or prediction confidences for optimizing on task-specific datasets; 
    \item Assessment of the performance on textual and multimodal visual QA tasks, demonstrating gains of up to +5.19\% with \ourname \ and up to +31.63\% with \ournameplus compared to ensembled base models on four benchmark datasets (Section \ref{main_results});
    \item A detailed analysis of data efficiency shows that \ourname~is more data efficient than state-of-the-art baseline ensemble methods. \ourname~surpasses the performance of the leading base models with as few as 10 training samples on one benchmark dataset  (Section \ref{main_results});
    \item An in-depth exploration of the importance of utilizing diverse information (image, question, answers) as fused inputs to enhance the performance of ensemble training (Section \ref{information_fusion} and \ref{fusion_or_concatenation}).\footnote{Code and data are available at \url{https://github.com/Yuuxii/Black-box-QA-Ensemling}.} 
\end{itemize}






\section{Related Work}

\textbf{Domain Adaptation.} These methods aim to improve the performance of a model on a task-specific domain by leveraging knowledge from other domains~\cite{zhou2022domain}. Methods such as fine-tuning~\cite{yosinski2014transferable}, feature adaptation~\cite{long2015learning} and data augmentation~\cite{choi2019self} aim to improve the performance of standalone models and thus typically require large amounts of labeled training data or access to the model architecture and weights. 

\textbf{Ensemble Learning.} Ensemble methods generate and combine multiple learners (ML models) to address a particular ML task~\cite{sagi2018ensemble}. Classical ensembling approaches like boosting~\cite{schapire2013explaining} and bagging~\cite{breiman1996bagging} are designed to train and combine a large number of individual models 
and are thus computationally expensive. 
Stacking~\cite{wolpert1992stacked} uses a meta-learner to integrate the probabilities of the predictions from base models for the final output. 
\citet{puerto2021metaqa} introduce MetaQA to select the best answer from multiple experts which requires both the knowledge of confidence score and base models' training data. Other methods train their base models to avoid dataset biases~\cite{han2021greedy}, while~\citet{xu2019open} aim to learn joint feature embeddings across different domains. However, these methods require at least one piece of knowledge that the black-box models can not provide, including base models' confidence scores or even training data (not available for ChatGPT). 

\textbf{Black-box Models Ensembling.} Black-box ensembling can be achieved by classical dynamic classifier selection methods, most notably OLA~\cite{woods1997combination}, which learns to rank the best classifier dynamically by its overall local accuracy in the nearest region of the input. Alternatively, majority voting~\cite{chan2022synthetic} and PageRank~\cite{Brin2012ReprintOT} weight the predictions by their internal agreements.
Yet, these methods are not designed for task-specific optimization and do not consider the information about inputs and predicted answers from base models simultaneously. 
To address this, \ourname~proposes a transformer-based setup that utilizes all the information available in the black-box setting, but not more, to enhance task-specific performance. 


\section{\ourname~ Ensemble Training}
Figure \ref{Fig:infosel} illustrates the proposed \ourname \ and \ournameplus \ frameworks to ensemble LLMs for TQA tasks (left), and VQA models for VQA tasks (right). 
We differentiate TQA components using LLMs and VQA components using VQA models by denoting them with superscripts $l$ and $v$ respectively. 
Similarly, to distinguish models used in TQA and VQA tasks, we add suffixes ``-TT'' and ``-MT'' respectively. For example, the \ourname-MT model in Figure \ref{Fig:infosel}, refers to the \ourname~for the VQA task.

Before training \ourname, we first perform the \textit{data preparation} (top of Figure \ref{Fig:infosel}) for both training and testing. Next, we train \ourname \ (step \raisebox{.5pt}{\textcircled{\raisebox{-.9pt} {1}}}). Training FT models (step \raisebox{.5pt}{\textcircled{\raisebox{-.9pt} {2}}}) and \ournameplus~(step \raisebox{.5pt}{\textcircled{\raisebox{-.9pt} {3}}}) 
is optional, but can significantly improve performance for datasets that contain a high percentage of labels that the base models have not been exposed to. 

\subsection{ InfoSel Training for TQA}

\textbf{Data Preparation.} 
First, we randomly sample $N$
content-question pairs $\{(C_i, Q_i)\}^{N}_{i=1}$ and the corresponding ground truth answers $\{\widetilde{A}^l_i \}^N_{i=1}$ from 
various benchmark datasets (refer to Section \ref{m:datasets}). 
Next, we build prompts $P_i$ following specific prompt rules $P_i = R(C_i, Q_i)$ (refer to Section \ref{m:datasets}). 
Using these prompts instead of plain $(C_i, Q_i)$ text improves the LLMs' answer quality \cite{bach2022promptsource}.
We select $K$ ($K$=3) black-box LLMs $\{M^l_j\}^K_{j=1}$  to generate answers on the $N$ prompts. The answer generated by $M^l_j$ on $P_i$ is denoted as $A^l_{ij}$ ($A^l_{ij} = M^l_j(P_i) $). Thereby, $K$ LLMs provide $N*K$ candidate answers for $N$ prompts. We calculate the word-level $F1$-scores \cite{rajpurkar2018know} of all the candidate answers $\{A^l_{ij}\}^K_{j=1}$ respectively for $P_i$. These $F1$-scores serve as target 
$ Y^l_i$ to optimize the ensemble model: 
\begin{equation}
Y^l_i = \{F1(A^l_{ij}, \widetilde{A}^l_i)\}^K_{j=1},  Y^l_i \in \mathbb{R} ^ {K}.
\end{equation}

The input for the ensemble training consists of $K$ texts. Each text is a joint string of $P_i$ and an answer predicted by a base model $j$, $A^l_{ij}$. 
More formally, the input $X^l_{i}$ is:
\begin{equation}
X^l_{i} = \{[P_i, A^l_{ij}]\}^K_{j=1}, \vert X_i^l \vert = K.
\end{equation}

The inputs $\{X^l_{i}\}_{i=1}^N$ and the corresponding target labels $ \{Y^l_i\}_{i=1}^N$ are used for ensemble training.

\textbf{\ourname-TT. } We use a textual BERT-base \cite{devlin2019bert} transformer $f^t_\theta$, ($\theta$ denote trainable model parameters) as the backbone of \ourname-TT. To achieve faster convergence, we load the pre-trained weights of \textit{bert-base-uncased} model. The input vector $X^l_{i}$ is passed to $f^t_\theta$ to generate $K$ sentence representations for each value in $X^l_{i}$  respectively. Thus, the sentence representation $R^{t}_{ij}$ of $[P_i, A^l_{ij}]$ from $f^t_\theta$ is:
\begin{equation}
R^{t}_{ij} = f^t_\theta([P_i, A^l_{ij}]), R^{t}_{ij} \in \mathbb{R} ^ {768}.
\end{equation}

A dense layer ($f^d_\theta$) is followed to classify $\{R^{t}_{ij}\}^K_{j=1}$, and is trained to match the target label $Y^l_i$ using binary cross entropy loss $\mathcal{L}_{BCE}$. More formally, the training objective of \ourname-TT is:
\begin{equation}
\underset{\theta}{\mathrm{min}} \sum_{i=1}^{N}\mathcal{L}_{BCE}(f^d_\theta([f^t_\theta([P_i, A^l_{ij}])]^{K}_{j=1}), Y^l_i).
\end{equation}

Finally, the trained \ourname-TT model ($M^{it}$) selects the winner model $M^l_{i, win}$ from $\{M_j^l\}^K_{j=1}$ for the input $P_i$ with the highest probability score based on the selection logits produced by  $f^d_\theta$. $A^l_{i, win}$ denotes the answer provided by $M^l_{i, win}$. 
\begin{equation}
\begin{aligned}
&M^l_{i, win} = M^{it}(\{[P_i, A^l_{ij}]\}^K_{j=1})\text{, } \\
&A^l_{i, win} = M^l_{i, win}(P_i).
\end{aligned}
\end{equation}

\textbf{FT-TT. } 
Using only the \ourname-TT model may limit performance due to the base models' lack of exposure to new (unseen) labels in the task-specific datasets.
To address this, we fine-tune a separate lightweight TT model directly on the TQA datasets to learn these new labels. 
Specifically, the training objective is to locate the start and end token position of the answer from the context $C_i$. 
We provide the 
token positions of $\widetilde{A}^l_i$ as the target label, such that the model is optimized to classify each token in two classes (start/end token). This fine-tuned textual transformer model is referred to as FT-TT ($M^{ft}$).\footnote{The training scheme is adapted from \url{https://huggingface.co/learn/nlp-course/chapter7/7?fw=pt} with the additional option to allow the model to return empty answers for unanswerable questions.} We denote the answer predicted by $M^{ft}$ on $P_i$ as $A^{ft}_i$.

\textbf{\ournameplus-TT. } 
This model performs a further ensemble training of FT-TT  and \ourname-TT models with the same training scheme and labeled training data as \ourname-TT. 
We anticipate that the thus trained \ournameplus-TT model on the output of \ourname-TT and the 
label finetuned FT-TT, will improve the ability to handle labels unseen by base models. As a result, we expect an improvement in the overall task-specific performance. The winner model selected by \ournameplus-TT belong to $\{M^{it}, M^{ft}\}$.

\subsection{ InfoSel  Training for VQA}

\textbf{Data Preparation.} Given $N$ image-question pairs $(I_i, Q_i)$ from dev data of VQA benchmark datasets, we use $K$ ($K$=3) 
pre-trained VQA models to predict answers $A^v_{ij}$ as follows: $\{M^v_j((I_i, Q_i))\rightarrow A^v_{ij}\}^K_{j=1}$.
We denote the ground truth answer for an image-question pair $(I_i, Q_i)$ as $\widetilde{A}^v_i$. Target labels $Y^v_i$ for ensemble training are given by the accuracy scores of the $K$ candidate answers evaluated on $\widetilde{A}^v_i$:\begin{equation}
Y^v_i = \{Acc(A^v_{ij}, \widetilde{A}^v_i)\}^{K}_{j=1} \text{, } Y^v_i \in \mathbb{R} ^ {K}.
\end{equation}

The joint string of a question ($Q_i$) with each of the candidate answers ($A^v_{ij}$) obtained from the base models, and the corresponding image ($I_i$) serve the input to our ensemble model \ourname-MT: 
\begin{equation}
X_i^v = \{(I_i, [Q_i, A^v_{ij}])\}^{K}_{j=1}, 
\vert X_i^v \vert = K.
\end{equation}

\textbf{\ourname-MT.} 
A Multimodal Transformer (MT, $f^m_\theta$)~\cite{li_unsupervised_2021} is employed as the backbone for \ourname-MT. Specifically, we first generate visual features $V_i$ of $I_i$ using a pre-trained R-CNN model ($M_r$)~\cite{anderson2018bottom}. $V_i$ is composed of a vector of the image region features $v_i$ and the detected \textit{tags} ( i.e., object labels of the image)~\cite{li_unsupervised_2021}. 
The joint string of question-answer pair $[Q_i, A^v_{ij}]$ and $V_i$ is then passed together to MT ($f_{\theta}^m$) to generate a fused contextual representation $R^m_{ij}$:
\begin{equation}
\begin{aligned}
&V_i = (v_i, tags) = M^r(I_i),\\
&R^{m}_{ij} = f^m_\theta(V_i, [Q_i, A^v_{ij}]), R^{m}_{ij} \in \mathbb{R} ^ {768}.
\end{aligned}
\end{equation}

Finally, we use an additional dense layer ($f_\theta^d$) to map $R^m_{ij}$ to the target label $Y^v_i$. 
The training is optimized using binary cross-entropy loss:
\begin{equation}
 \underset{\theta}{\mathrm{min}}\sum_{i=1}^{N}\mathcal{L}_{BCE}(f^d_\theta([f^m_\theta(V_i, [Q_i, A^v_{ij}])]^{K}_{j=1}), Y^v_i).
\end{equation}

We denote $M^{im}$ to the trained \ourname-MT model, $M^{im}$  selects the winner model $M^v_{i, win}$ from $\{M_j^v\}^K_{j=1}$ to predict answer $A^v_{i, win}$ for the image-question pair $(I_i, Q_i)$ 
based on the selection logits produced by $f^d_\theta$.
\begin{equation}
\begin{aligned}
&M^v_{i, win} = M^{im}(\{(I_i, [Q_i, A^v_{ij}])\}^K_{j=1})\text{, } \\
&A^v_{i, win} = M^v_{i, win}(I_i, Q_i).
\end{aligned}
\end{equation}

\textbf{FT-MT.} Similar to FT-TT, FT-MT composed of a trainable MT and a Multilayer Perceptron (MLP) is fine-tuned with the same training data as \ourname-MT. Differently, FT-MT solves a multi-label classification task by classifying the fused contextual representation of  $Q_i$  (instead of $[Q_i, A^v_{ij}]$ like \ourname-MT) and $V_i$ to a predefined answer list (labels). This list contains frequent answers from the training data. As a result, a trained FT-MT model ($M^{fm}$) can learn to predict the unseen (new) labels (answers) contained in the 
task-specific datasets, but not in the pre-training data of the base models. $A_i^{fm}$ denotes the answer predicted by $M^{fm}$ over $(I_i, Q_i)$. The training scheme is adapted from ~\cite{li_unsupervised_2021}.

\textbf{\ournameplus-MT. } Similar to \ournameplus-TT,  \ournameplus-MT model ensembles the FT-MT and \ourname-MT models using the same training scheme as in \ourname-MT. The winner model selected by 
\ournameplus-MT belong to $\{M^{im}, M^{fm}\}$.

\section{Experiments and Analysis}

\subsection{Datasets} \label{m:datasets}
To demonstrate the \textit{data efficiency} of our approach, we subsampled four publicly available benchmark datasets. his process resulted in the creation of four \textit{Mini} datasets. The training set comprises less than 1\% of the original size of the TQA datasets and about 10\% of the original size of the original VQA datasets.
Table \ref{tab:datadetails} presents the statistics of these datasets.

\begin{table}[h]
\centering
\renewcommand{\arraystretch}{1.25}
\caption{Details of the \textit{Mini} datasets used for \ourname \/ ensemble training and testing. \% stands for the percentage of the original full dataset.}
\label{tab:datadetails}
\begin{tabular}{l l c c  }
\toprule
\textbf{\textit{Mini} Dataset}& \textbf{Source Dataset} & \textbf{Num.} & \textbf{\%} \\
\hline
{Mini-SDv2 train} & SQuAD-V2 train&  800 &0.56\\
{Mini-SDv2 val}& SQuAD-V2 train & 200 & 0.14\\
{Mini-SDv2 test} &  SQuAD-V2 dev & 11,873 & 8.39\\
\hline
{Mini-NQ train} & NQ-Open train & 800 &0.87 \\

{Mini-NQ val} & NQ-Open train & 200 & 0.22\\
{Mini-NQ test} & NQ-Open dev & 3,499 & 3.83 \\
\hline
{Mini-GQA train} & GQA dev & 105,640 &9.80\\
{Mini-GQA val} & GQA dev  & 26,422 
 & 2.45\\
{Mini-GQA test} & GQA test & 12,578 & 1.17\\
\hline
{Mini-Viz train} & VizWiz dev & 3,456 &10.5 \\
{Mini-Viz val} & VizWiz dev & 863 & 2.63\\
{Mini-Viz test} & VizWiz test & 8,000 & 24.39\\

\bottomrule
\end{tabular}

\end{table}

\subsubsection{TQA Datasets} 

\textbf{SQuAD-V2~\cite{rajpurkar2018know}} stands for Stanford Question Answering Dataset 2.0, a dataset designed for the task of question answering. It is an extension of the original SQuAD dataset by including over 50,000 unanswerable questions written adversarially by crowdworkers. The dataset is widely used in natural language understanding research. 

\textbf{NQ-Open~\cite{kwiatkowski-etal-2019-natural}} is derived from Natural Questions and serves as an open-domain question-answering evaluation. The entirety of the questions can be addressed using the information found in the English Wikipedia. It was created for research purposes.

We generated two \textit{Mini} datasets, Mini-SDv2 and Mini-NQ, consisting of 1,000 randomly sampled instances from SQuAD-V2~\cite{rajpurkar2018know} and NQ-Open~\cite{kwiatkowski-etal-2019-natural} train splits, respectively.

For Mini-NQ, we followed~\citet{fisch2019mrqa} to use long answers as the context, and short answers as the ground truth answers. The 1,000 samples are divided into train and validation data using an 8:2 ratio, while the trained models are tested on the dev data of the original datasets due to the unavailability of original test data. We use the setup proposed in~\cite{laskar-etal-2023-systematic} to generate prompts with PromptSource~\cite{bach2022promptsource} for the Mini-SDv2 dataset. These prompts integrated with context and questions from the dataset are used to query answers from LLMs. The prompts for Mini-NQ, unavailable in PromptSource, are adapted from the prompts of Mini-SDv2.  The details of these prompts are shown in Table \ref{tab:prompt}.  



\begin{table*}
\centering
\caption{Prompts in TQA datasets. SQuAD-V2 is available in PromptSource~\cite{bach2022promptsource} for prompt generation, we use the prompts from PromptSource for Mini-SDv2, which contains two forms of prompts. NQ-Open dataset is not available in PromptSource and thus the prompt for Mini-NQ is adapted from Mini-SDv2.}
\renewcommand{\arraystretch}{1.25}
\label{tab:prompt}
\begin{tabular*}{\tblwidth}{l@{\hspace{0.5cm}} p{14.8cm} }
\toprule
\textbf{Dataset} &  \textbf{Sample Prompts} \\ \hline

 \multirow{4}{*}{Mini-SDv2} &  What is the answer? {\textbackslash}n Context:[context]; {\textbackslash}n Question:[question]; {\textbackslash}n If you can't find the answer, please respond "unanswerable" {\textbackslash}n  Answer: \\
\cline{2-2}
 &  Answer the question depending on the context. {\textbackslash}n Context: [context]; {\textbackslash}n Question: [question]; {\textbackslash}n If you can't find the answer, please respond "unanswerable". {\textbackslash}n Answer:  \\
\hline
  \multirow{2}{*}{Mini-NQ} & Answer the question depending on the context without explanation. {\textbackslash}n Context: [context]; {\textbackslash}n Question: [question]; {\textbackslash}n Answer: \\
\bottomrule
\end{tabular*}

\end{table*}


\subsubsection{VQA Datasets} \label{vqa-dataset}

\textbf{GQA~\cite{hudson2019gqa}}  is a large-scale dataset for visual reasoning and compositional question answering research. The dataset contains over 113k images collected from a diverse set of sources and over 22 million questions. Only one ground-truth answer is provided for each image-question pair.

\textbf{VizWiz~\cite{gurari2018vizwiz}}  is a benchmark dataset for visual question answering. It includes 31K images, 250K questions, and answers collected through a mobile app for visually impaired users. 10 ground-truth answers are provided for each image-question pair. 

\begin{figure}[t!]
\centering
\caption{Top 7 most frequent answers of VQA v2 (pre-trained dataset of VQA base models), GQA and VizWiz (task-specific datasets for ensemble training). This explains why \ourname \ performs poorly in Mini-Viz, as the new label "unanswerable" is the top frequent answer in VizWiz, but this new label has been exposed to base models that are pre-trained on VQA v2. }
\label{fig:data}
\includegraphics[width=\linewidth]{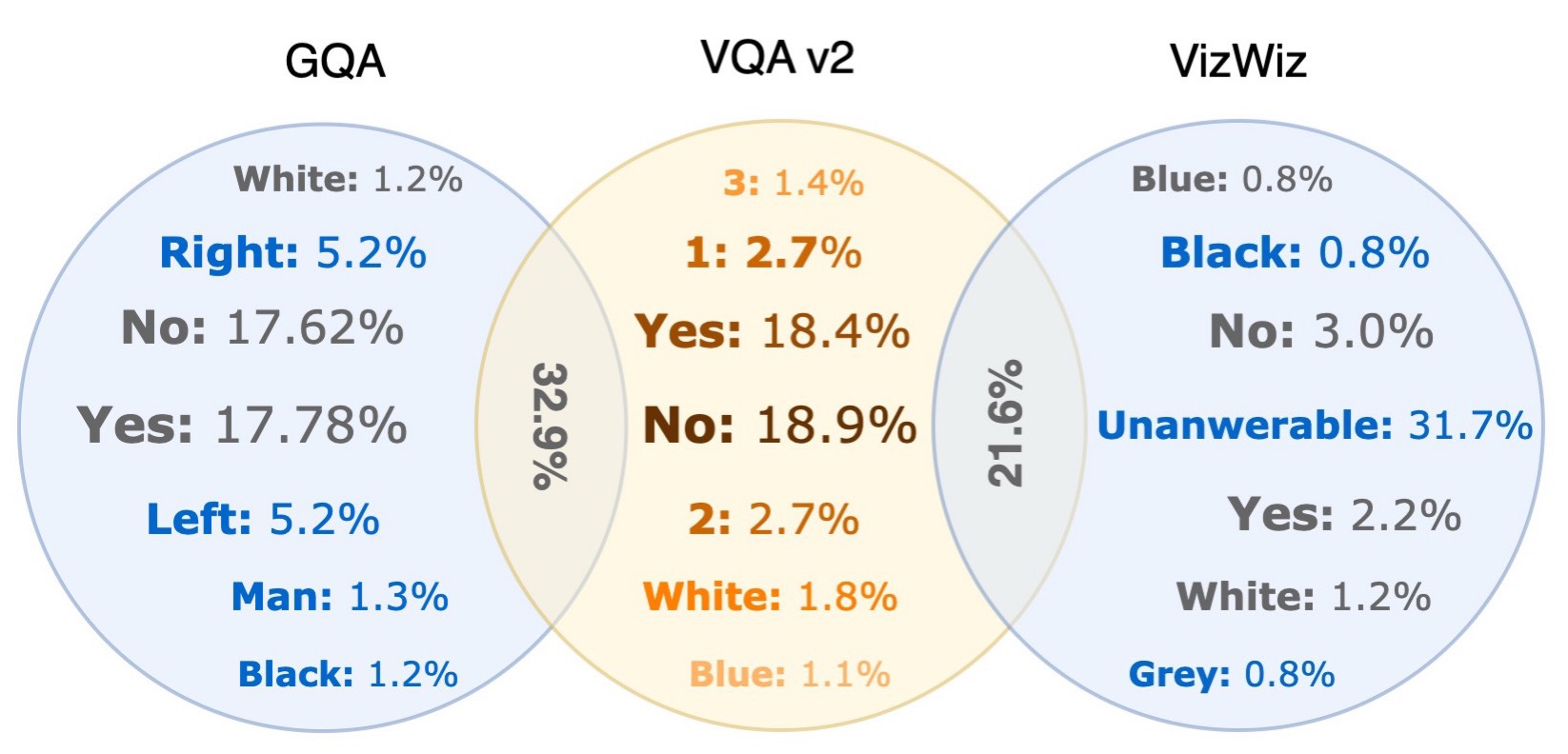}
\end{figure}

Mini-GQA and Mini-Viz are sampled from GQA~\cite{hudson2019gqa} and VizWiz~\cite{gurari2018vizwiz}. As the training data of the VQA models are publicly accessible in contrast to LLMs, we compare the label differences of the pre-trained dataset (VQA v2~\cite{antol2015vqa}) of VQA base models with task-specific datasets (GQA, VizWiz) for ensemble training. Figure \ref{fig:data} shows the top 7 most frequent answers and their percentages in GQA, VQA v2 and VizWiz. Four answers in GQA do not appear in the top list of VQA v2 and three for VizWiz (including the top frequent answer "unanswerable").  We also sampled 3k of the most frequent answers from each dataset and calculated their percentage of overlap, which is reported on the intersection in the figure. GQA and VizWiz have 32.9 \% and 21.6\% of overlap with VQA v2 respectively, showcasing significant differences between the pre-trained dataset and task-specific datasets.  

Our results (Figure \ref{Fig:qa-per}) reveal that VQA tasks demand a greater quantity of training samples compared to TQA tasks.
Therefore, we constructed the train and validation sets of Mini-GQA and Mini-Viz datasets using a larger fraction (the dev data) of GQA~\cite{hudson2019gqa} and VizWiz~\cite{gurari2018vizwiz} datasets. Similar to the TQA datasets, the train and validation split ratio is 8:2 ratio.

\subsection{Baselines}

We use Majority Voting (MV), Weighted Voting (WV)~\cite{schick2020exploiting}, PageRank~\cite{Brin2012ReprintOT},  Overall Local Accuracy (OLA)~\cite{woods1997combination}, PairRanker and LLM-Blender~\cite{jiang2023llmblender} as baselines. 

\textbf{Majority Voting (MV).} MV makes a collective decision by considering the predicted answers as a group of individuals voting on a particular input. The answer that receives the most votes is the winner, otherwise, ties are broken randomly.

\textbf{Weighted Voting (WV).} 
We adopt a strategy similar to ~\citet{schick2020exploiting}, where the model accuracy of the train data before training is used as the weight for average weighting. In our case, we use the corresponding accuracy of the base models as the weight for voting.

\textbf{PageRank}~\cite{Brin2012ReprintOT}. We adapt PageRank as a baseline to determine the most suitable answer in a graph where all the answers to one question are connected by their BLEURT~\cite{sellam2020bleurt} similarities.

\textbf{Overall Local Accuracy (OLA)}~\cite{woods1997combination}. Following~\cite{cruz2018dynamic}, we employ the k-nearest neighbors algorithm to divide the input space (comprising representations of prompts for TQA, of images and of questions for VQA) of training data into 7 regions. The overall local accuracy of each base model in different regions is computed as its region competence. The model presenting the highest competence level is selected to predict the label of inputs that fall in the same region.

\textbf{PairRanker} ~\cite{jiang2023llmblender}. The  PairRanker model is trained to rank a pair of candidate predictions from two LLMs using multiple optimizing objectives (i.e., EM, F1). In the end, the pairwise rank probability produced by the PairRanker model is used to select the top 1 answer as the final prediction. We train PairRanker using the same textual transformer backbone as \ourname \ for comparison.

\textbf{LLM-Blender}~\cite{jiang2023llmblender}. 
 LLM-Blender is the extension of PairRanker, which feed the top k (we set k to 2 in the experiments) predictions from the PairRanker model to a GENFUSER model (Flan-T5-XL, as described in \cite{chung2022scaling}, which has 3 billion parameters) to generate the final fused prediction. 
We use pre-trained (0-shot) GENFUSER model from the original paper which have been trained over massive data (105k) including TQA data to test on our data.

\subsection{Base Models}
We experiment with ensembling GPT-3.5-turbo-0613 (ChatGPT), LLaMA-2-70b-chat (hereinafter referred to as ``LLaMA'')~\cite{touvron2023llama} and GPT-3.5 text-davinci-003 (hereinafter referred to as ``Davinci'') to generate answers for TQA tasks. To tackle VQA tasks, we employ three VQA models (VLMo~\cite{bao2022vlmo}, ALBEF~\cite{li_align_2021} and BLIP~\cite{li_blip_2022}), which are pre-trained on VQA v2 dataset~\cite{antol2015vqa}. Note that we use publicly accessible VQA models to save experimental costs, but these models are assumed to be black-box. 

\begin{table}[h!]
\centering
\caption{Parameter size of the models used in our experiments.}
\label{tab:model_param}
\begin{tabular*}{\tblwidth}{c@{\hspace{1.5cm}} l@{\hspace{1.5cm}} c }
\toprule
& \textbf{Model } &  \textbf{\#Param.}\\\hline
\multirow{7}{*}{TQA} & {LLaMA-2-70b-chat} &{70B} \\
& {text-davinci-003}& {175B} \\
& {ChatGPT} &  {175B} \\

& FT-BERT & 110M \\
& PairRanker& {110M}\\
& \ourname-TT& {110M}\\
& \ournameplus-TT& {110M}\\

\hline
\multirow{7}{*}{VQA}& {ALBEF} &  {290M}\\
& {BLIP}& {361M} \\
& {VLMo} &{182M}\\
& FT-MT & 115M\\
& PairRanker& {110M}\\
& {\ourname-MT}& {115M}\\
& {\ournameplus-MT}& {115M}\\
\bottomrule

\end{tabular*}

\end{table}

\textbf{ChatGPT} is a large language model with 175B parameters, it allows you to have human-like conversations and much more with the chatbot. ChatGPT can generate context-based responses to user prompts (questions). However, this model is currently only accessible by cloud APIs.

\textbf{GPT 3.5 text-davinci-003} is similar to ChatGPT but designed specifically for instruction-following tasks which enables it to respond concisely and more accurately. This model was deprecated for public access after our experiments, but this did not influence the effectiveness of our method.

\textbf{LLaMA-2-70b-chat~\cite{touvron2023llama}} is a large language model with 70B parameters. It is fine-tuned on LLaMA 2 with publicly available instruction datasets and over 1 million human annotations, while Llama 2 models are trained on 2 trillion tokens from publicly available online data sources. LLaMA 2 models are currently publicly accessible.

\textbf{ALBEF~\cite{li_align_2021}}\footnote{\url{https://github.com/salesforce/ALBEF}} first encodes the image and text with an image encoder (visual transformer~\cite{dosovitskiy2020image}) and a text encoder respectively. Then a multimodal encoder is used to fuse the image features with the text features through cross-modal attention. The V\&L representation is trained with objectives of image-text contrastive learning, masked language modeling and image-text matching. Unlike U-VisualBERT, ALBEF uses a 6-layer transformer decoder to generate answers for VQA tasks. 

\textbf{BLIP~\cite{li_blip_2022}}\footnote{\url{https://github.com/salesforce/BLIP}} utilizes a visual transformer as the image encoder, and a multi-task model (multimodal mixture of encoder-decoder) as a unified model with both understanding and generation capabilities.  The model is jointly pre-trained with three vision-language objectives: image-text contrastive learning, image-text matching, and image-conditioned language modeling. Similarly to ALBEF, this method treats the VQA task as an answer generation task.

\textbf{VLMo~\cite{bao2022vlmo}}\footnote{\url{https://github.com/microsoft/unilm/tree/master/vlmo}} is a unified vision-language pre-training method with Mixture-of-Modality-Experts. VLMO leverages large-scale image and text data to learn joint representations of vision and language. It employs a mixture model to capture diverse interactions between visual and textual information, achieving state-of-the-art performance on various vision-language tasks.

Table \ref{tab:model_param} reports the parameter size of the base models and their ensemble models (\ourname-TT and \ourname-MT). 
\ourname~ provides an efficient method for ensembling large LLMs such as ChatGPT (175B parameters) using only 110M trainable parameters.
Even though only  37\% ((182M-115M)/182M) trainable parameters are saved for the VQA task, we still demonstrate that \ourname~can effectively enhance the task-specific performance of small-size black-box VQA models, offering a lightweight solution. 

\subsection{Evaluation Metric} 
LLMs tend to generate contextual answers that lead to lower scores in the exact match (\textbf{EM}).
Therefore, we mainly use the (per-answer) token-level \textbf{F1}-score from the official evaluation guidance of the datasets as the main evaluation metric for TQA performance. Our results differ from the ones reported in~\cite{laskar-etal-2023-systematic, Koco__2023} because we do not apply any post-processing, human evaluation or output constraints on the generated answers. 

Finally, we use \textbf{Oracle} to represent the maximum capability of a combination of base models. Specifically, for each input, the Oracle selects the answer with the highest agreement to the ground truth among all the answers predicted by the base models. 
Thus, the Oracle score represents the performance of an ideal ensemble model.


\begin{table*}[ht!]
\centering
\renewcommand{\arraystretch}{1.25}
\caption{\label{tab:llm_main} Test performance comparison on TQA datasets. The overall best results are highlighted in bold, and the best results of base models are underlined. \textit{Input Info.} stands for the fused input information required by different models/methods to make predictions. $Q$ denotes questions, and $A$ denotes the answers generated by base models.}

\begin{tabular}{ll@{\hspace{.5cm}}c@{\hspace{1cm}}c@{\hspace{1cm}}c@{\hspace{1cm}}c@{\hspace{1cm}}c@{\hspace{1cm}}c@{\hspace{0.5cm}}}
\toprule
\noalign{\vskip 0.02cm}
& \multirow{2}{*}{\textbf{Model/Method}} & \multicolumn{2}{c@{\hspace{1cm}}}{\textbf{Input Info.}} &\multicolumn{2}{c@{\hspace{1cm}}}{\textbf{Mini-SDv2}} & \multicolumn{2}{c@{\hspace{0.5cm}}}{\textbf{Mini-NQ}}\\
\cline{3-8}
& & $Q$ & $A$ &  \textbf{EM} &\textbf{F1}& \textbf{EM} & \textbf{F1} \\
\noalign{\vskip 0.02cm}
\hline
\noalign{\vskip 0.02cm}
& LLaMA-2-70b-chat & \checkmarkicon{lightblue}{white} & \xmark & 0.24  & 11.34    & 28.07 &  46.47\\

\textbf{Base} &text-davinci-003 & \checkmarkicon{lightblue}{white} & \xmark & \underline{52.37} & \underline{58.44} & 52.24 &  69.44\\
\textbf{Models} & ChatGPT &	\checkmarkicon{lightblue}{white} & \xmark & 30.89  & 44.95  & \underline{57.53} & \underline{71.54}\\
& \textbf{Oracle} & - & - & 58.61  & 66.20& 64.02 &  79.21\\
\noalign{\vskip 0.02cm}
\hline  
\hline
\noalign{\vskip 0.02cm}
& MV & \xmark & \checkmarkicon{lightblue}{white} & 26.95 &  37.75 & 46.07  & 62.43\\
& WV  & \xmark& \checkmarkicon{lightblue}{white}&  52.37 &  58.44 & 57.53 & 71.54\\

 & PageRank \cite{Brin2012ReprintOT} & \xmark & \checkmarkicon{lightblue}{white}& 25.39 &37.31 & 51.76 &68.53\\
\textbf{Baselines} & FT-BERT \cite{yosinski2014transferable} & \checkmarkicon{lightblue}{white}& \checkmarkicon{lightblue}{white} & 46.80 & 47.68& 36.52  & 40.60\\
& OLA \cite{woods1997combination} & \checkmarkicon{lightblue}{white}& \checkmarkicon{lightblue}{white}& 47.90 &55.59 & 54.70 & 70.05\\
& PairRanker \cite{jiang2023llmblender} & \checkmarkicon{lightblue}{white}& \checkmarkicon{lightblue}{white}& 57.28 & 63.33 & 57.96 & 72.21 \\
& (0-shot) LLM-Blender \cite{jiang2023llmblender} &\checkmarkicon{lightblue}{white} &\checkmarkicon{lightblue}{white} &4.90 &21.20 & 1.03& 25.06 \\

\noalign{\vskip 0.02cm}
\hline
\hline
\noalign{\vskip 0.02cm}


\multirow{4}{*}{\textbf{Ours}}&  & \checkmarkicon{lightblue}{white} &\xmark   & 57.50 & 63.60 & \textbf{59.07}  &\textbf{73.46} \\
& \textbf{\ourname-TT} & \xmark & \checkmarkicon{lightblue}{white}& 57.51  & 63.58 & 58.30  &72.93 \\
&  &\checkmarkicon{lightblue}{white} & \checkmarkicon{lightblue}{white}& \textbf{57.74}  & \textbf{63.71}&  58.45 & 73.37\\
\cline{2-8}
& \textbf{\ournameplus-TT } & \checkmarkicon{lightblue}{white}& \checkmarkicon{lightblue}{white}&  49.09  & 49.85 & 48.16 & 53.70 \\
\noalign{\vskip 0.02cm}	
\bottomrule

\end{tabular}

\end{table*}

\begin{figure*}
\caption{TQA test performance of  \ourname~compared to baselines over increasing size of training data. \textit{Best base} represented the best performance of the base models.}\label{Fig:qa-per}
\centering
   \begin{minipage}{0.45\textwidth}
     \centering
     \includegraphics[width=1.\linewidth]{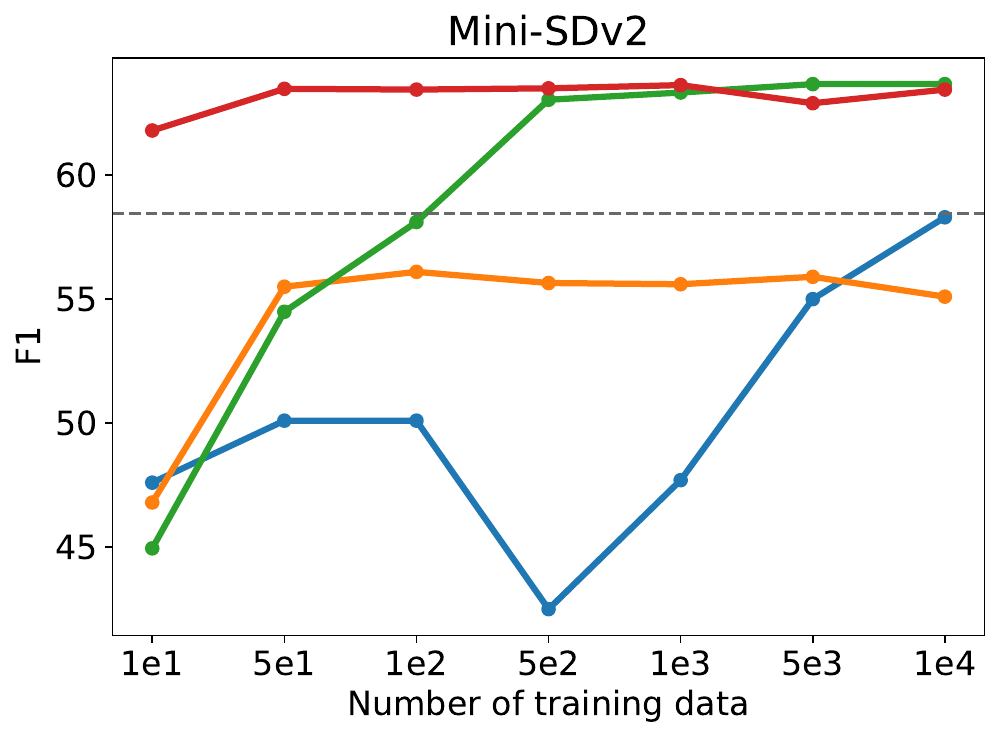}\\
   \end{minipage}
   \hspace{0.2cm} 
   \begin{minipage}{0.45\textwidth}
     \centering
     \includegraphics[width=1.\linewidth]{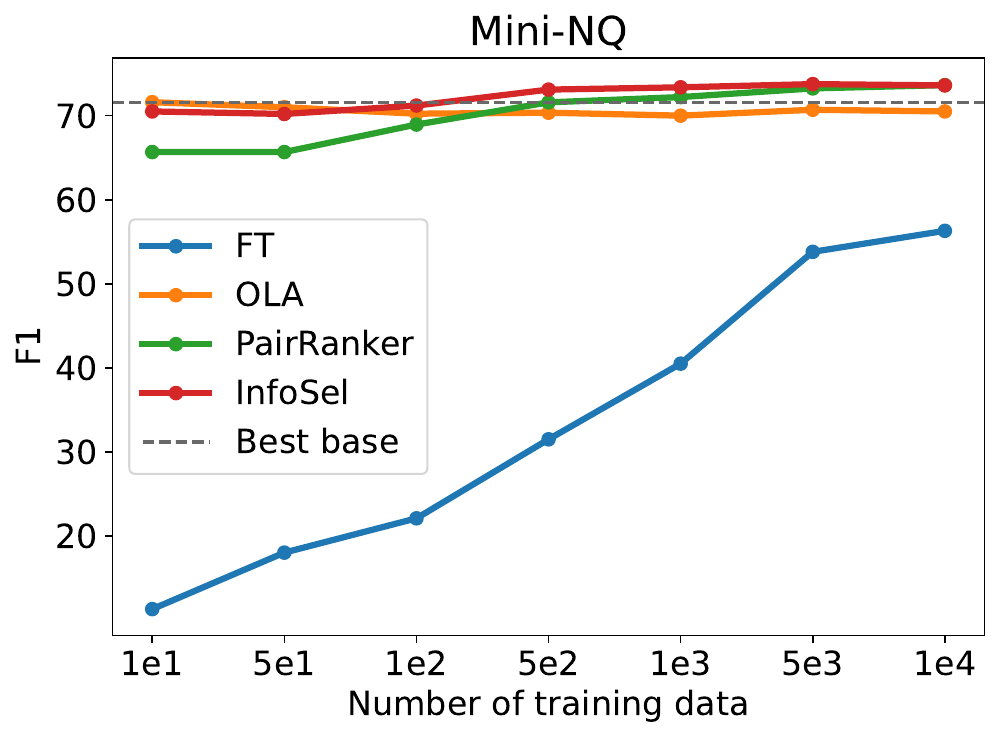}\\
   \end{minipage}

\end{figure*}

\begin{table*}[ht!]
\centering
\renewcommand{\arraystretch}{1.25}
\caption{\label{tab:vqa_main} Test performance comparison on VQA datasets. The test annotation of Mini-Viz dataset is not accessible and thus the oracle score can not be reported. The overall best results are highlighted in bold, and the best results of base models are underlined. \textit{Input Info.} stands for the fused input information required by different models/methods to make predictions. $V$ and $Q$ denote visual and question information, $A$ denote the answers generated by the base models.}
\begin{tabular}{ll@{\hspace{.5cm}}c@{\hspace{1cm}}c@{\hspace{1cm}}c@{\hspace{1cm}}c@{\hspace{1cm}}c@{\hspace{.5cm}}}
\toprule
&\multirow{2}{*}{\textbf{Model/Method}}& \multicolumn{3}{c@{\hspace{1cm}}}{\textbf{Input Info.}}& \textbf{Mini-GQA} & \textbf{Mini-Viz}\\
\noalign{\vskip 0.02cm}

\cline{3-7}
\noalign{\vskip 0.02cm}
& & $V$ & $Q$ & $A$ &\multicolumn{2}{c@{\hspace{0.5cm}}}{\textbf{ACC}}\\
\hline
\noalign{\vskip 0.02cm}
& ALBEF& \checkmarkicon{lightblue}{white} & \checkmarkicon{lightblue}{white} & \xmark &  \underline{50.60} & \underline{21.28}\\
\textbf{Base} & BLIP & \checkmarkicon{lightblue}{white} & \checkmarkicon{lightblue}{white} & \xmark & 48.08 &20.80\\
\textbf{Models} & VLMo &\checkmarkicon{lightblue}{white} & \checkmarkicon{lightblue}{white} & \xmark &	48.21 &19.77\\
& \textbf{Oracle} &- & - & \xmark &  65.03 & - \\

\noalign{\vskip 0.02cm}
\hline 
\hline
\noalign{\vskip 0.02cm}

& MV & \xmark & \xmark & \checkmarkicon{lightblue}{white} & 51.05&21.47\\
& WV  & \xmark & \xmark & \checkmarkicon{lightblue}{white} & 52.10&19.43\\
& PageRank \cite{Brin2012ReprintOT} &\xmark & \xmark & \checkmarkicon{lightblue}{white} &51.47&21.66\\
\textbf{Baselines} & FT-MT \cite{yosinski2014transferable} &\checkmarkicon{lightblue}{white} & \checkmarkicon{lightblue}{white}& \checkmarkicon{lightblue}{white} & 50.48& 51.76\\
& OLA \cite{woods1997combination} &\checkmarkicon{lightblue}{white} & \checkmarkicon{lightblue}{white}& \checkmarkicon{lightblue}{white} &48.65&20.32 \\

& PairRank \cite{jiang2023llmblender}  &\xmark & \checkmarkicon{lightblue}{white}& \checkmarkicon{lightblue}{white}& 52.05 & 22.42\\
& (0-shot) LLM-Blender \cite{jiang2023llmblender} & \xmark & \checkmarkicon{lightblue}{white}& \checkmarkicon{lightblue}{white}&0.0 &0.0 \\
\noalign{\vskip 0.02cm}
\hline
\hline
\noalign{\vskip 0.02cm}

\multirow{9}{*}{\textbf{Ours}}& \multirow{8}{*}{\textbf{\ourname-MT}}  &\checkmarkicon{lightblue}{white} & \xmark & \xmark & 50.56 &20.79 \\
&  &\xmark & \checkmarkicon{lightblue}{white}& \xmark & 51.11 & 21.21 \\
&  &\xmark & \xmark & \checkmarkicon{lightblue}{white}& 51.69& 22.83 \\
&  &\checkmarkicon{lightblue}{white} & \checkmarkicon{lightblue}{white}& \xmark& 50.83 & 20.06 \\
&  &\checkmarkicon{lightblue}{white} & \xmark & \checkmarkicon{lightblue}{white}& 52.38 & 22.66 \\
&  & \xmark & \checkmarkicon{lightblue}{white}& \checkmarkicon{lightblue}{white}&54.76 & 22.89 \\
&  & \checkmarkicon{lightblue}{white} & \checkmarkicon{lightblue}{white}& \checkmarkicon{lightblue}{white}&\textbf{55.16}&23.16\\
\cline{2-7}
& \textbf{\ournameplus-MT} &\checkmarkicon{lightblue}{white} & \checkmarkicon{lightblue}{white}& \checkmarkicon{lightblue}{white}& 52.54&\textbf{52.91}\\
	
\noalign{\vskip 0.02cm}	
\bottomrule

\end{tabular} 
\end{table*}





\subsection{Experiment Setup}
We fixed the batch size to the upper limit of the server capacity, while the learning rates and epochs are selected after a grid search on a set of values (learning rates: \{e3, 5e4, e4, 5e5, e5, 5e6, e6\}, epochs: \{3, 5, 10, 15, 20\}).  Models for TQA are trained for 5 epochs using a learning rate of $5\times10^{-5}$ and batch size of 4. 
 Models for VQA use the same learning rate but a batch size of 16 for 20 epochs. We spent $\sim$74 and $\sim$290 seconds training 1 epoch on 1,000 samples for TQA and 4,319 samples for VQA respectively. The training was performed on 1 GPU with 16GB memory of a DGX1 server ((Pascal) Tesla P100). 


\begin{figure*}[ht!]
   \centering
   \caption{VQA test performance of  \ourname~compared to baselines over increasing size of training data. \textit{Best base} represented the best performance of the base models. The FT method is outperforming all the ensemble methods (OLA, PairRanker and \ourname) due to the underexposure of a new label "unanswerable" to the ensembled base models.}\label{Fig:vqa-per}
   \begin{minipage}{0.45\textwidth}
     \centering
     \includegraphics[width=1.\linewidth]{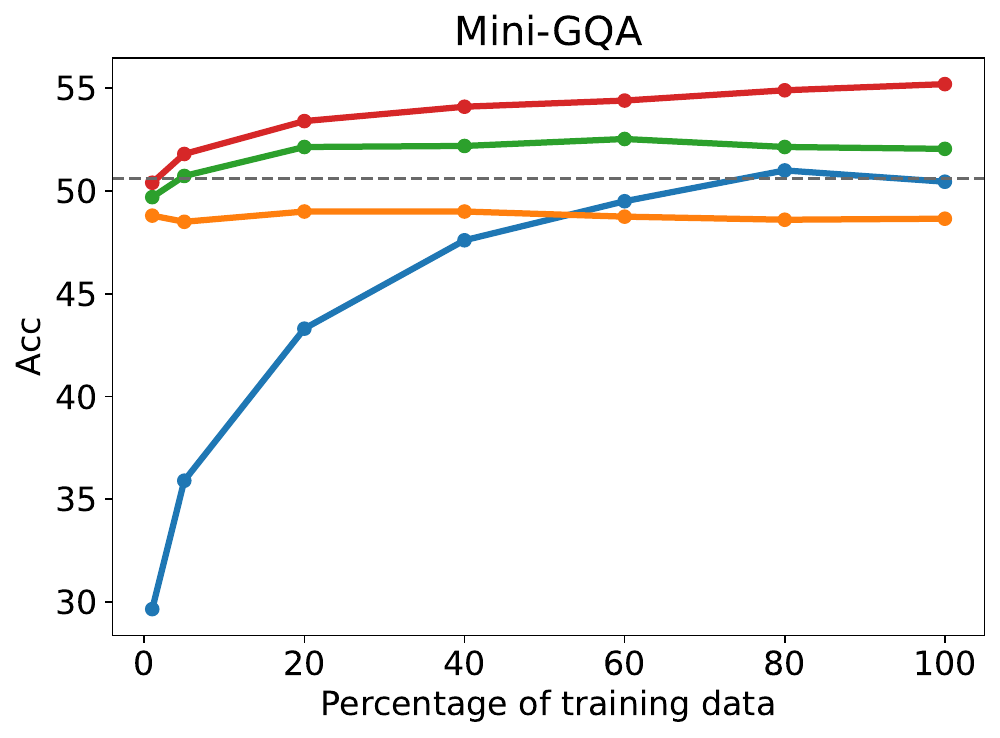}\\
   \end{minipage}
   \hspace{0.2cm} 
   \begin{minipage}{0.45\textwidth}
     \centering
     \includegraphics[width=1.\linewidth]{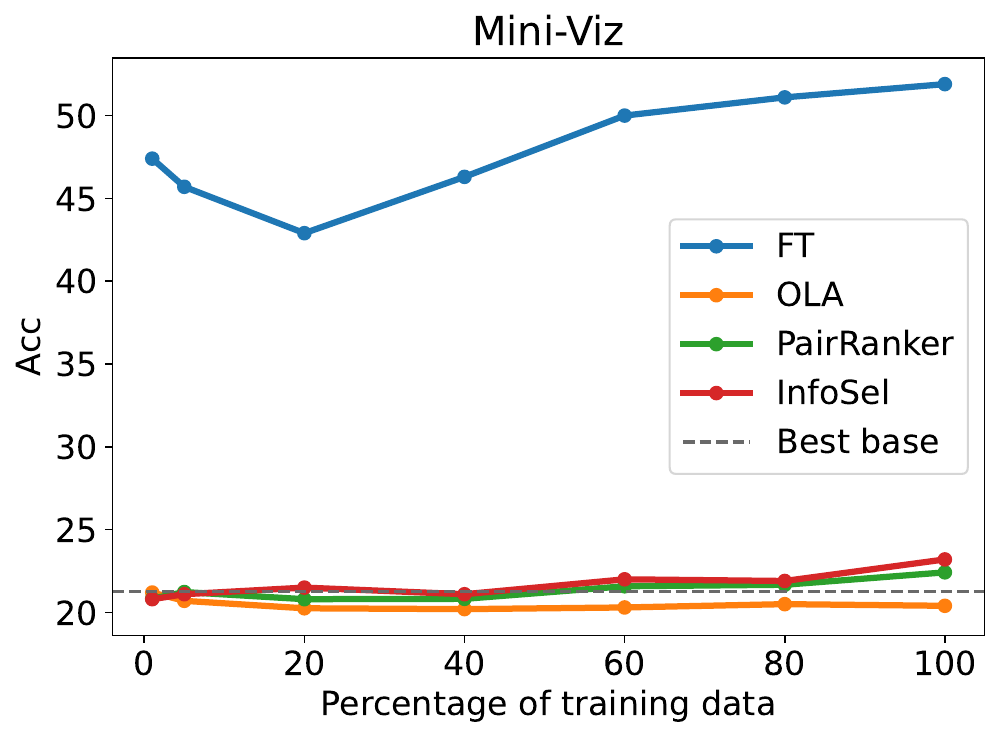}\\
   \end{minipage}
   
\end{figure*}

\section{Results and Analysis}

In this section, we analyze the performance of our method, taking into account its distinctive characteristics as described in Table \ref{tab:compare}. Concretely, we focus on comparing our models in terms of \textit{task-specific} performance, \textit{data efficiency}, and \textit{multimodal} capabilities.
\subsection{Main Results} \label{main_results}
\textbf{Task-specific Performance.} Table \ref{tab:llm_main} and \ref{tab:vqa_main} demonstrate the  \textit{task-specific} performance of \ourname, base models and ensemble baselines on textual and visual QA datasets. For TQA, we observe that LLaMA underperforms other base models. Upon closer examination, we found that LLaMA generates longer explanation text which, although often accurate, decreases the EM and F1-score values. Conversely, a more consistent performance of base models is observed for VQA. 
Mini-Viz contains 28\% of unanswerable questions, and the label ``unanswerable'' has never been seen by base models. Consequently, this lack of exposure leads to significantly lower performance scores. 

 Baseline ensemble methods such as WV, PageRank and OLA achieve only marginal improvements compared to base models ($\le$+1.5\%) on VQA datasets. 
 These results highlight the limitations of these methods when applied to \textit{task-specific} datasets (see also Table \ref{tab:compare}).  
 PairRanker \ underperformed \ourname \ even though it has been trained with the same data. 0-shot LLM-Blender tends to generate longer answers compared to PairRanker which leads to a low score especially when evaluated in the exact-match settings (EM, ACC). 

 
For TQA task, \ourname-TT achieves 5.19\% (63.63-58.44) improvement compared to individual base models, and reaches 96.12\% (63.63/66.20) of the Oracle on Mini-SDv2. Similarly, the corresponding improvement in Mini-NQ performance is 1.83\%, reaching 93.06\% of the performance achieved by the Oracle. 
In contrast, FT-TT, despite its superior performance over two base models on Mini-SDv2, underperforms \ourname-TT by more than 15\% due to the small-size training data (refer to Figure \ref{Fig:qa-per}). 
This low performance of FT-TT models 
negatively impacts the performance of \ournameplus. As a result, we conclude that while \ournameplus~can exhibit superior performance (see further for VQA task), it also requires more training data. 

For VQA task, the results in Table \ref{tab:llm_main} showcase an improvement of 4.56\% in the accuracy score of \ourname-MT compared to the base models (55.16-50.60) on Mini-GQA.
Furthermore, FT-MT 
improves 30.48\% (51.76-21.28) 
accuracy on Mini-Viz due to a high percentage of unseen labels (e.g., ``unanswerable'') introduced during fine-tuning. 
Finally, the superior performance of \ournameplus-MT model on Mini-Viz dataset demonstrates the effectiveness of the proposed blending approach, which improves 31.63\% (52.91-21.28) accuracy upon the \ourname-MT model. 

\textbf{Data Efficiency.} 
The experimental results shown in Figure \ref{Fig:qa-per} and \ref{Fig:vqa-per} demonstrate the \textit{data efficiency} of our method by evaluating the performance of the models across varying training data sizes.
We observe that \ourname-TT achieves a higher F1-score compared to the best base model when trained on as little as 10 samples from Mini-SDv2. This result has been further verified with the mean F1-score of 10 test results using different seed variations for sampling the training data (shown in Figure \ref{Fig:sd}). Conversely, the number of training samples needed to surpass the performance of base models is higher for VQA datasets: 5\% (6,603 samples) for Mini-GQA and 20\% (864 samples)  for Mini-Viz. We hypothesize that this is due to the inherent complexity of the VQA task. 

PairRanker is less data-efficient than \ourname \ as it only achieves close performance when the training samples are more than 500 on both Mini-SDv2 and Mini-NQ. Additionally, we find that a larger training data size benefits FT-TT more than \ourname-TT and OLA. For example, the F1-score of FT-TT increases  $\sim$200\% and $\sim$500\% from 10 to 10,000 training samples on Mini-SDv2 and Mini-NQ respectively, while \ourname-TT shows only marginal improvements of $\sim$3\% and $\sim$4\%. However, FT-TT still underperforms the best base model,  which suggests that fine-tuning a small-size model requires larger training data for getting a comparable performance with LLMs or \ourname. 
Finally, we observe that a larger training data size 
does not necessarily lead to improved performance for the fine-tuned FT-TT model (e.g., when increasing from 80\% to 100\% the training data size on Mini-GQA).  
In contrast, OLA does not benefit as much as \ourname~and FT from a larger size of training data, only outperforming \ourname-TT on Mini-NQ when utilizing 10 and 20 training samples. 

\textbf{Multimodal Data.} \ourname~is able to utilize multimodal data (image and text) for VQA tasks, and thus outperform the latest text-exclusive LLM ensemble methods (PairRanker and LLM-Blender) as evidenced in Table \ref{tab:llm_main}. In contrast, PairRanker and LLM-Blender cannot process image features, thereby lacking crucial information in the multimodal setting, leading to a low accuracy on VQA datasets.\footnote{The most frequent answer of LLM-Blender on VQA datasets is \emph{``I'm sorry, I don't have enough context to answer that question.''}}
Further insights into the significance of modality information are elaborated in Section \ref{information_fusion}.

\subsection{Importance of Different Input Information Fused for Ensembling} \label{information_fusion}
In this section, we analyze what input information is important to fuse to improve ensembling performance. 

For results in Table \ref{tab:llm_main}, we observe that all \ourname-TT models using different kinds of input information surpass the base models and baselines. The performance differences in using different input information are quiet margin (<0.7\%).  The best performance is achieved by using both the question and answer as fused input information for \ourname-TT on Mini-SDv2. Differently, the question information works the best training on Mini-NQ. 

Table \ref{tab:vqa_main} presents the performance results of the \ourname-MT models utilizing various kinds of input information during the ensemble training process. Our model outperforms PairRanker when using the same kind of input information - namely, the question and answer. Additionally, incorporating the answer information alongside any supplementary data (whether visual or textual) enables our model to achieve superior performance compared to baseline models and existing benchmarks.
The setting that yields the lowest accuracy solely utilizes the image (V) as the signal. This lower performance can be attributed to the fact that a single image often corresponds to multiple questions in VQA datasets, making it challenging for the model to acquire discriminative features. Furthermore, we conclude that the superior performance of our model when utilizing image, question, and answer (V+Q+A) data demonstrates the effectiveness of our model in \textit{multimodal} setting.

\subsection{Input Information Concatenation or Fusion} \label{fusion_or_concatenation}

We studied the impact of concatenating and fusing multi-modal input information for the VQA task. \ourname-MLP is an alternative model type for \ourname\ which processes all the input information separately with a simple Multilayer perceptron (MLP) instead of MT. A pre-trained Sentence-BERT ~\cite{reimers2019sentence}
\footnote{\url{https://huggingface.co/sentence-transformers/multi-qa-mpnet-base-dot-v1}}  $M_{qa}$ is used for generating question embedding $R^q$ and answer embeddings $R^a$.
\begin{equation}
\begin{aligned}
&R_i^q = M^{qa}(Q_i), R^q \in \mathbb{R} ^ {768}, \\
&R^{a}_{ij} = M^{qa}(A^v_{ij}), R^{a}_{ij} \in \mathbb{R} ^ {768}  
\end{aligned}
\end{equation}
MLP takes the concatenated representation of question, answer, and visual embeddings $R^v_{i}$ as input and maps it to the label space. The objective function of \ourname-MLP is formalized as:
\begin{equation}
    \underset{\theta}{\mathrm{min}} \sum_{i=1}^{N}\mathcal{L}_{BCE}(MLP_\theta(\{R_i^v, [R_i^{q}, R^a_{ij}]\}^K_{j=1}), Y^v_i)
\end{equation}

The input layer of the MLP maps the concatenated representations to a hidden layer with a dimension equal to 300, followed by a ReLU activation layer and then an output layer with an output size equal to the number of models. 

Table \ref{tab:mlp} presents the performance comparison between the input concatenation method (\ourname-MLP) and the fusion method (\ourname-MT). Our results indicate that \ourname-MT achieves approximately 3\% higher accuracy on the Mini-GQA dataset and about 2\% higher accuracy on the Mini-Viz dataset compared to \ourname-MLP. This suggests that the fused contextual representation of inputs provides more discriminative information than simply concatenating input embeddings.

\begin{table}
\centering
\renewcommand{\arraystretch}{1.25}
\caption{Comparison of using different architecture for processing input information differently. Input concatenation result is demonstrated by \ourname-MLP and the fusion result is shown by \ourname-MT.}
\label{tab:mlp}
\begin{tabular*}{\tblwidth}{l@{\hspace{.8cm}}c@{\hspace{.8cm}} c@{\hspace{.8cm}}c@{\hspace{.2cm}}}
\toprule
\multirow{2}{*}{\textbf{Model}} & \textbf{Input} &\textbf{Mini-GQA} & \textbf{Mini-Viz}\\
\cline{3-4}
 & \textbf{Fuse}& \multicolumn{2}{c@{\hspace{.2cm}}}{\textbf{ACC} }\\

\hline
\ourname-MLP & \xmark &52.35&21.12\\
\textbf{\ourname-MT} & \checkmarkicon{lightblue}{white} & \textbf{55.16}&\textbf{23.16}\\	
\bottomrule
\end{tabular*} 

\end{table}

\subsection{Ablation Studies}\label{ablation}

\begin{figure*}[ht!]
   \centering
\caption{The portions of answers selected from different base models by \ourname~models on Mini-SDv2 and Mini-NQ test data.   }\label{Fig:model_selection}
\includegraphics[width=.75\linewidth]{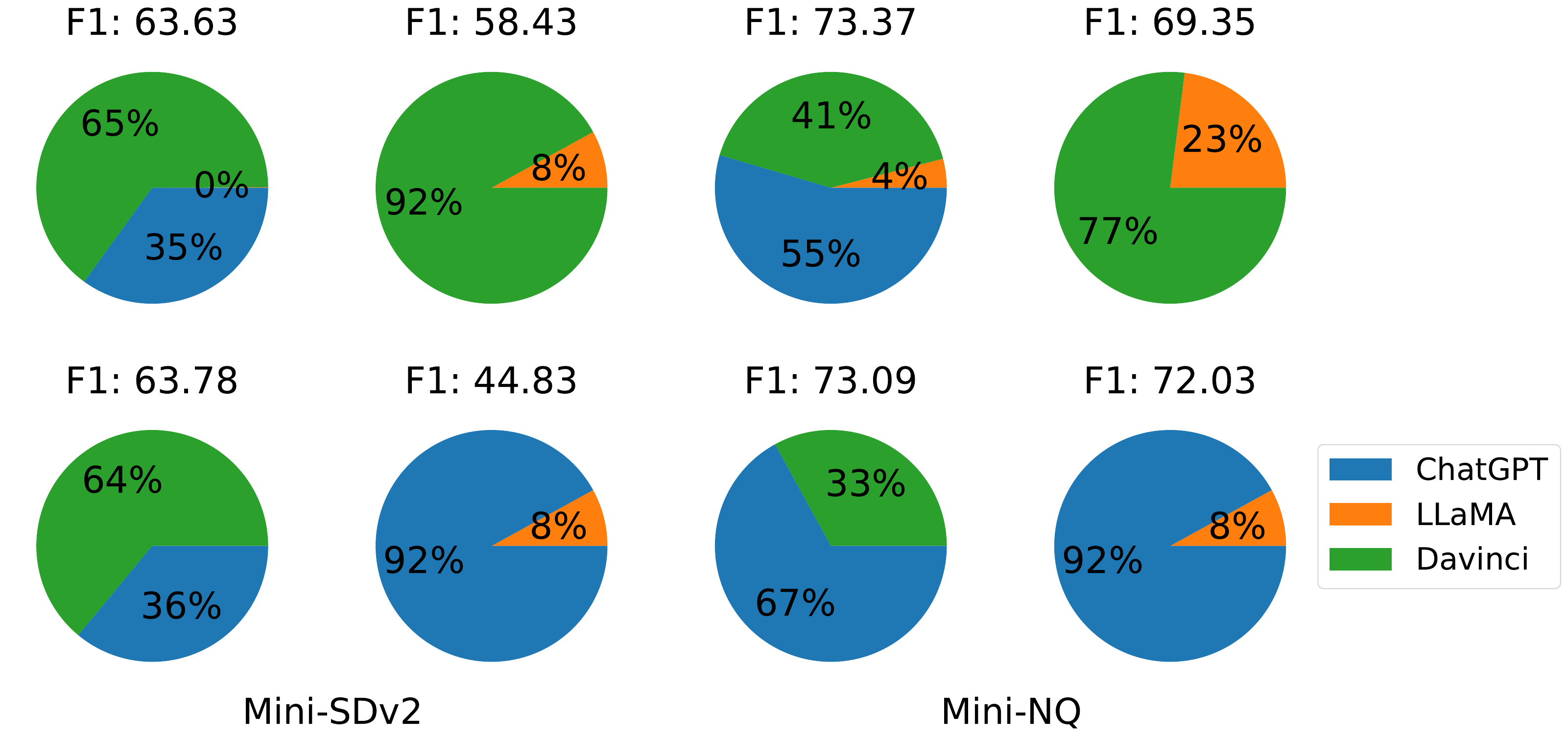}\\
 
\end{figure*}


\textbf{Is \ourname~robust to the base models' individual performances?}
We carry out this study to assess whether \ourname \ can effectively utilize the predictions obtained from various base models, regardless of their individual performances.
In the first column of each dataset in Figure \ref{Fig:model_selection}, we observe a minor F1-score difference (0.15\%) on the Mini-SDv2 dataset between the \ourname~model ensembled with and without the lowest performing base model (LLaMA). This finding suggests that \ourname~is robust, and not significantly affected by the performance of individual models.  In a more detailed analysis, we observe that \ourname~selects 4\% of answers from LLaMA on Mini-NQ, resulting in an overall gain of +0.28\% of the F1-score. This observation highlights the effectiveness of \ourname, as it can leverage the knowledge contained in the answers provided even by the lowest performing base model to some extent. However, the results of the second column of each dataset reveal that the ensemble performance decreased significantly when the ChatGPT or Davinci model was removed. This indicates that \ourname~ should not be used to ensemble two base models (e.g., LLaMA and ChatGPT) with large performance gaps.

\subsection{Case Studies}

\begin{table*}[t!]
\centering
\caption{\label{tab:qacase} Case study of our models on Mini-SDv2 test and Mini-NQ test data. Answers of LLMs are shortened to keywords for better demonstration. Ground-truth answers are bolded, and one suspicious ground-truth answer is colored red.
}
\renewcommand{\arraystretch}{1.25}
\begin{tabular*}{\tblwidth}{lcccc}
\toprule

 &  \multicolumn{2}{c}{\textbf{mini-SDv2}} &  \multicolumn{2}{c}{\textbf{mini-NQ}} \\
 
 \multirow{4}{*}{\textbf{Context:}} & ... The building was& ... Derrick Norman   &  Dwight David    & ... in 2005 and the\\
  \textbf{} & designed by architects& Lehmer's list   &  Howard ... player  &  release of her epony- \\
  \textbf{} & Marek Budzyński and&of primes up to    & for the Charlotte  &  mous debut album\\
 \textbf{} & Zbigniew Badowski...& 10,006,721 ... & Hornets ...  &  the following year ...\\
  \hline
\multirow{4}{*}{\textbf{Question:}}  & What profession  & How many primes were &who did Dwight   & when did Taylor  \\
 & does Zbigniew    &  included in Derrick  & Howard play  & Swift 's first \\
  \textbf{} & Marek have?  &  Norman Lehmer's list    & for last year?& album release?\\
   \textbf{} &  &  of prime numbers?  & & \\
\hline

LLaMA-2-70b-chat  &architect& unanswerable & \textbf{Charlotte Hornets} & 2006\\
text-davinci-003  & Architect& \textbf{10,006,721} & The Houston Rockets & 2006\\
ChatGPT &\textbf{unanswerable} & unanswerable& Washington Wizards &2006\\
FT-BERT & architects Marek Budzyński  & unanswerable & Dwight David Howard &\textbf{\textcolor{red}{2005}}\\

 & and Zbigniew Badowski & \textbf{} &  &\\
\hline
\textbf{\ourname-MT} & \textbf{unanswerable} &\textbf{10,006,721} & \textbf{Charlotte Hornets}&  {2006}\\
\textbf{\ournameplus-MT} & \textbf{unanswerable} &unanswerable & Dwight David Howard& \textbf{\textcolor{red}{2005}}\\

\bottomrule

\end{tabular*}

\end{table*}

\begin{table*}[t!]
\centering
\caption{\label{tab:vqacase} Case study of our models on Mini-GQA test and Mini-Viz validation data. Ground-truth answers are bolded.
}
\renewcommand{\arraystretch}{1.25}
\begin{tabular*}{\tblwidth}{l@{\hspace{1cm}}c@{\hspace{1.2cm}}c@{\hspace{1.2cm}}c@{\hspace{1.2cm}}c@{\hspace{1.2cm}}}
\toprule

 &  \multicolumn{2}{c@{\hspace{1.2cm}}}{\textbf{Mini-GQA}} &  \multicolumn{2}{c@{\hspace{1.2cm}}}{\textbf{Mini-Viz}} \\
 \textbf{Image:} & \includegraphics[width=0.16\linewidth, height=20mm]{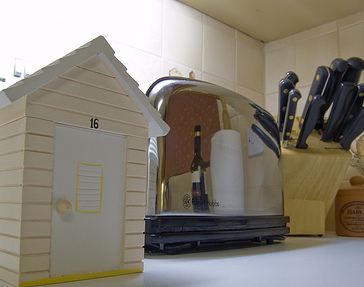}  & \includegraphics[width=0.15\linewidth, height=20mm]{images/n46510.jpg} & \includegraphics[width=0.11\linewidth, height=20mm]{images/VizWiz_val_00003569.jpg} & \includegraphics[width=0.10\linewidth, height=20mm]{images/VizWiz_val_00003469.jpg}\\

  \multirow{2}{*}{\textbf{Question:}} & What appliance is    &Is the tall tree on & What kind of food& What is this pro-  \\

 \textbf{} & it?& the right? & is in this can? & duct?\\
\hline
ALBEF &blender& yes & fruit salad & refrigerator\\
BLIP  & \textbf{toaster} & yes & \textbf{vegetable soup} & toilet\\
VLMo &microwave & yes& fruit &door\\
FT-MT & coffee maker & \textbf{no} & soup &\textbf{unanswerable}\\
\hline
\textbf{\ourname-MT} & \textbf{toaster} &yes& \textbf{vegetable soup}& toilet\\
\textbf{\ournameplus-MT} & coffee maker &\textbf{no} & \textbf{vegetable soup}& \textbf{unanswerable}\\

\bottomrule

\end{tabular*}

\end{table*}



Table \ref{tab:qacase} and \ref{tab:vqacase} showcase several interesting cases from the predictions of different models for textual and visual QA tasks. From Table \ref{tab:qacase}, we observe that that \ourname-TT selects answers from different language models. However, \ournameplus-TT may select incorrect answers from the overfitted FT-BERT model , leading to underperformance compared to \ourname-TT in those instances. The final case illustrates a incorrect ground-truth answer provided by the original dataset. Despite this, LLMs generate the correct answer due to their contextual comprehension capabilities. In contrast, FT-BERT, which is limited to classification tasks, can only extract answer tokens from the context and thus fails to provide the correct answer. Therefore, we conclude that leveraging an ensemble of LLMs to harness their robust comprehension abilities offers greater benefits to users than relying on fine-tuning smaller models.

Table \ref{tab:vqacase} demonstrates that both \ourname-MT and \ournameplus-MT are capable of identifying the correct answer, even when the prediction of only one of the base models is accurate. The last case demonstrates that \ournameplus-MT successfully identifies the new label "unanswerable" introduced by FT-MT, a label that \ourname-MT cannot predict because the base models can not provide. Consequently, incorporating FT-MT into the ensembling training process is crucial when out-of-domain datasets contain a high proportion of new labels.

\subsection{Robustness Analysis} 
\begin{figure}[t!]
     \centering
     \caption{Mean and standard deviation of results on Mini-SDv2 and Mini-NQ when training \ourname-TT with 10 sets of randomly sampled training data.}\label{Fig:sd}
     \includegraphics[width=.95\linewidth]{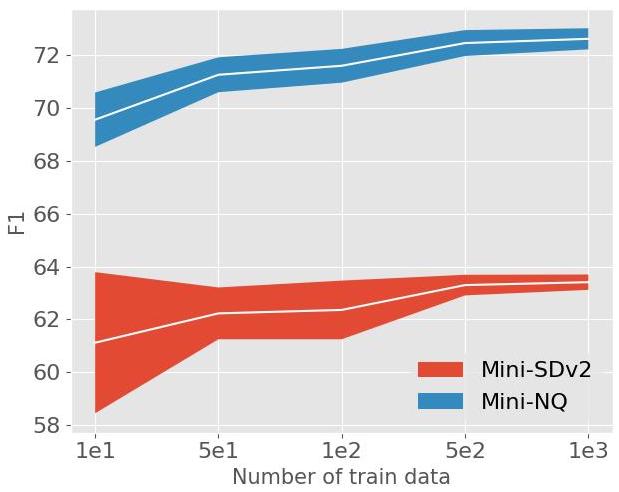}
\end{figure}

We analyze the robustness of the data efficiency property of \ourname~by training \ourname-TT with 10 different sets of randomly sampled train data. Figure \ref{Fig:sd} demonstrates the mean and standard deviation of these 10 sets of results on Mini-SDv2 and Mini-NQ. We observe that the deviation decreases as the number of training samples increases. Concretely, when using 1,000 training samples, the deviation is less than 0.5 (0.29 for Mini-SDv2, 0.40 for Mini-NQ). Remarkably, the mean F1-score (61.12) achieved with just 10 samples from Mini-SDv2 surpasses that of the best base model. Furthermore, the mean F1-score (71.60) obtained with 100 samples from Mini-NQ exceeds the F1-score of the best base model, despite using fewer samples than the 500 samples reported in the main result figure (Figure \ref{Fig:qa-per}).

\subsection{Discussion and Limitations}

\ourname\ offers an effective approach to enhancing out-domain black-box model performance and addressing answer selection. However, it is important to acknowledge certain limitations associated with its application, which are outlined in the following paragraphs.

Dependency on Annotated Data: \ourname, similar to numerous machine learning techniques, relies on a small amount of annotated training and development data tailored to a specific domain. While this requirement is relatively modest, and the efficiency of \ourname~in data utilization has been demonstrated through experiments, it may still pose a limitation in scenarios where obtaining such data is challenging or costly. 

Limited Applicability to Open-Ended Text Generation: the primary strength of \ourname lies in its ability to select the best answer from a set of base models, making it particularly valuable in question-answering scenarios. However, for more open-ended text-generation tasks, where it may be beneficial to combine multiple answers, \ourname's single-answer selection mechanism may not be the ideal choice, and future research directions may include approaches for combining several long-form answers.


Transparency and Explainability: \ourname, similar to other machine learning models that derive answers from black-box systems, may also function as a "black box". This implies that its decision-making process may not be easily interpretable or understandable to end-users. By integrating \ourname~with explainability techniques, users may gain a more transparent view of how the model arrives at its selections.

\section{Conclusion}

In this paper, we propose \ourname{}, a novel lightweight and task-specific ensemble method designed to learn the dynamic selection of the optimal answer from a range of distinct black-box base models.  \ourname~is able to handle multimodal data and we demonstrate its effectiveness on textual and visual question answering tasks. 

Our findings also show that \ourname is highly data-efficient and, using only
a fraction of the trainable parameters relative to the base models, 
our method can substantially increase the performance over the best-performing base LLMs and VQA models. 
Concretely, it requires only a fraction of instances from the training set to outperform base LLMs and previous ensembling methods.
An in-depth exploration of the importance of different input information for ensembling shows that fused (rather than concatenated) information of diverse types (image, question, answer) achieved the best performance. An ablation and robustness analysis reveals that \ourname{} 
is robust to incorrect predictions of low-performing LLMs and to the randomness of training data selection.


\printcredits

\section*{Declaration of competing interest}
The authors declare that they have no known competing financial interests or personal relationships that could have appeared to influence the work reported in this paper.

\section*{Data availability}
The data used in this paper is publicly accessible.

\section*{Acknowledgements}
This research has been funded by the Vienna Science and Technology Fund (WWTF) [10.47379/VRG19008] “Knowledge-infused Deep Learning for Natural Language Processing”. Co-funded by the European Union (grant Agreement no. 101146515).\footnote{Views and opinions expressed are however those of the authors only and do not necessarily reflect those of the European Union or European Research Executive Agency. Neither the European Union nor the granting authority can be held responsible for them.}

\bibliographystyle{cas-model2-names}

\bibliography{custom, anthology}



\end{document}